\documentclass[12pt]{l4dc2020} 

\title[Analysis of the Expressiveness of Deep Neural Network Architectures]{An Analysis of the Expressiveness of Deep Neural Network Architectures Based on Their Lipschitz Constants}
\usepackage{times}

\author{%
 \Name{Siqi Zhou} \Email{siqi.zhou@robotics.utias.utoronto.ca}%
 \AND
 \Name{Angela P. Schoellig} \Email{schoellig@utias.utoronto.ca}\\[0.5em]
 \addr Dynamic Systems Lab, Institute for Aerospace Studies, University of Toronto, Canada\\
 \addr Vector Institute for Artificial Intelligence, Canada
}
\newcommand{\Rbb}{\mathbb{R}}

\newcommand{\Ebb}{\mathbb{E}}

\newcommand{\bx}{\mathbf{x}}

\definecolor{grey1}{RGB}{192,192,192}
\definecolor{grey2}{RGB}{178,178,178}
\definecolor{grey3}{RGB}{150,150,150}
\definecolor{grey4}{RGB}{119,119,119}
\definecolor{grey5}{RGB}{77,77,77}
\definecolor{green}{RGB}{112,173,71}
\definecolor{blue2}{RGB}{68,115,196}
\definecolor{red}{RGB}{192,0,0}
\definecolor{yellow}{RGB}{255,192,0}

\usepackage{wrapfig}

\usepackage[font=small]{caption}

\newenvironment{justification}{\paragraph{Justification}}{\hfill\jmlrBlackBox}

\usepackage{bigints}

\begin{document}

\maketitle
\sloppy

\begin{abstract}%
Deep neural networks (DNNs) have emerged as a popular mathematical tool for function approximation due to their capability of modelling highly nonlinear functions. Their applications range from image classification and natural language processing to learning-based control. Despite their empirical successes, there is still a lack of theoretical understanding of the representative power of such deep architectures. In this work, we provide a theoretical analysis of the expressiveness of fully-connected, feedforward DNNs with 1-Lipschitz activation functions. In particular, we characterize the expressiveness of a DNN by its Lipchitz constant. By leveraging random matrix theory, we show that, given sufficiently large and randomly distributed weights, the expected upper and lower bounds of the Lipschitz constant of a DNN and hence their expressiveness increase exponentially with depth and polynomially with width, which gives rise to the benefit of the depth of DNN architectures for efficient function approximation. This observation is consistent with established results based on alternative expressiveness measures of DNNs. In contrast to most of the existing work, our analysis based on the Lipschitz properties of DNNs is applicable to a wider range of activation nonlinearities and potentially allows us to make sensible comparisons between the complexity of a DNN and the function to be approximated by the DNN. We consider this work to be a step towards understanding the expressive power of DNNs and towards designing appropriate deep architectures for practical applications such as system control.
\end{abstract}

\vspace{0.5em}

\section{Introduction}
\label{sec:introduction}
Given their capability to approximate highly nonlinear functions, deep neural networks (DNNs) have found increasing application in domains such as image classification~\cite{krizhevsky2012imagenet,googlenet}, natural language processing~\cite{hinton2012deep,hannun2014deep}, and learning-based control~\cite{shi2019neural,chen2019large,zhou-cdc17}. As compared to their shallow counterparts, DNNs are often favoured in practice due to their compact representation of nonlinear functions~\cite{montufar2017number}. Despite their practical successes, the theoretical understanding of the representative power of such deep architectures remains an active research topic addressed by both the machine learning and neuroscience community. In this work, we aim to contribute to the understanding of the expressiveness of DNNs by presenting a new perspective based on Lipschitz constant analysis that is interpretable for applications such as system control. 

There are several recent works analyzing the expressive power of deep architectures. One notable work is~\cite{NIPS2011_4350}, where the authors show that, for a sum-product network, a deep network is exponentially more efficient than a shallow network in representing the same function. Following this work, several researchers then considered more practical DNNs with piecewise linear activation functions (e.g., rectified linear units (ReLU) and hard tanh) and showed that the expressiveness of a DNN measured by the number linear regions partitioned by the DNN grows exponentially with depth and polynomially with width~\cite{pascanu2013number,montufar2014number,arora2016understanding,serra2017bounding}. 
In parallel to the work on piecewise linear DNNs, \citep{raghu2017expressive} consider DNNs with independent and identically distributed (i.i.d.) Gaussian weight and bias parameters (i.e., random DNNs) and introduce a new measure of expressiveness based on the length of the output trajectory as the DNN traverses a one-dimensional trajectory in its input space. Similar to the other results, the authors show that the expressiveness of a DNN measured by the expected output trajectory length increases exponentially with the depth of the network.

While existing work has shown the exponential expressiveness of deep architectures, the measures of expressiveness are typically specific to the type of deep architectures being considered. For instance, for the sum-product networks considered in~\cite{NIPS2011_4350}, the measure of expressiveness is the  number of monomials used to construct the polynomial function, and for DNNs with piecewise linear activation functions~\cite{pascanu2013number,montufar2014number,arora2016understanding,serra2017bounding}), the number of linear regions is used as the measure to characterize the complexity of the DNN. These specialized notions of expressivity prohibit sensible comparisons between the complexity of a DNN and the underlying function it approximates.  
While the expressiveness measure based on output trajectory length~\cite{raghu2017expressive} is applicable to DNNs with more general activation functions, it is still not trivial to connect this measure to the properties of the function to be approximated by the DNN.

In this work, motivated by the theoretical analysis of DNNs in feedback control applications~\cite{shi2019neural,fazlyab2019efficient}, we introduce an alternative perspective on the expressive power of DNNs based on their Lipschitz properties. 
Similar to~\cite{raghu2017expressive}, we consider a DNN with random weight parameters. By leveraging results from random matrix theory, we provide an analysis of the expressive power of DNNs based on their Lipschitz constant and establish connections with earlier results using alternative measures of DNN expressiveness. Our ultimate goal is to understand the implications of choosing particular neural network architectures for learning in feedback control applications.

\section{Preliminaries}
We consider fully-connected DNNs, $f: \mathcal{X}\mapsto \mathcal{Y}$, that are defined as follows:
\begin{equation}
\label{eqn:dnn_definition}
\begin{aligned}
\mathbf{h}_0(\mathbf{x}) = \bx,\:\:
\mathbf{h}_l(\bx) =\boldsymbol{\sigma}\left( \mathbf{W}_l\mathbf{h}_{l-1}(\bx)+\mathbf{b}_l\right)\: \forall l = 1,...,L,\:\:
\mathbf{y}=\mathbf{W}_{L+1}\mathbf{h}_{L}(\bx) + \mathbf{b}_{L+1},
\end{aligned}
\end{equation}
where $\bx\in \mathcal{X}\subseteq \Rbb^{n_0}$ is the input, $\mathbf{y}\in \mathcal{Y}\subseteq \Rbb^{n_{L+1}}$ is the output, the subscripts $l = \{0,...,L+1\}$ denote the layer index with $l=0$ being the input layer, $l=1,...,L$ being the hidden layers, and $l=L+1$ being the output layer, $\mathbf{h}_l:\mathcal{X} \mapsto \Rbb^{n_l}$ is the output from the $l$th layer with $\boldsymbol{\sigma}(\cdot)$ being the element-wise activation function and $n_l$ being the number of neurons in the $l$th layer, and $ \mathbf{W}_l\in \Rbb^{n_l\times n_{l-1}}$ and $\mathbf{b}_l\in\Rbb^{n_l}$ are the weight and bias parameters between layers $(l-1)$ and $l$. In our analysis, we focus on DNNs with 1-Lipschitz activation functions~\cite{virmaux2018lipschitz}, which include most commonly used activation functions such as ReLU, tanh, and sigmoid.

To facilitate our analysis, similar to~\cite{raghu2017expressive}, in this work, we consider DNNs with random weight matrices $\mathbf{W}_l$ whose elements are i.i.d. zero-mean Gaussian random variables~$\mathcal{N}(0,\sigma_w^2)$, where $\sigma_w^2$ is the variance of the Gaussian distribution. Our goal is to analyze the expressiveness of such a DNN as we vary its architectural properties (i.e., width and depth).

\section{Lipschitz Constant as a Measure of Expressiveness}
\label{sec:lipschitz_constant}
In this work, we characterize the expressiveness of a DNN by its Lipschitz constant. Intuitively, a larger Lipschitz constant implies that small changes in the DNN input can lead to large changes at the output, which provide greater flexibility to model nonlinear functions.

Formally, a function $f:\mathcal{X}\mapsto\mathcal{Y}$ is said to be Lipschitz continuous on $\mathcal{X}$ if
\begin{equation}
\label{eqn:lipschitz_constant_definition}
(\exists \rho >0)\:(\forall \bx,\bx'\in\mathcal{X}) \:\: ||f(\bx)-f(\bx')||\le \rho||\bx-\bx'||,
\end{equation}
and its Lipschitz constant on $\mathcal{X}$ is the smallest $\rho$ such that the inequality in~\eqref{eqn:lipschitz_constant_definition} holds. It is not hard to verify that common activation functions (e.g., ReLU, tanh, and sigmoid) are globally Lipschitz continuous. A DNN with such activation functions is a finite number of compositions of Lipschitz continuous functions and is thus Lipschitz continuous on its domain $\mathcal{X}$. Note that, in general, the Lipschitz continuity condition in~\eqref{eqn:lipschitz_constant_definition} is independent of the choice of the norm; in this work, we will consider Lipschitz continuity in the $l_2$-norm.

In the following subsections, we establish a connection between the expected Lipschitz constant of a DNN and its architecture (i.e., width and depth), and compare the result to existing results on the expressive power of DNNs in the literature. We summarize our main results in this manuscript and provide details of the derivations and proofs in the appendices.

\subsection{Upper and Lower Bounds on the Lipschitz Constant of a DNN}
\label{subsec:lipschitz_expressivness}
As noted in \cite{fazlyab2019efficient,virmaux2018lipschitz}, the exact estimation of the Lipschitz constant of a DNN is NP-hard; however, for our purpose of understanding the expressiveness of DNNs, estimates of the upper and lower bounds on the Lipschitz constant of a DNNs based on their weight matrices are sufficient.

Recall that we consider a family of DNNs with 1-Lipschitz activation functions. By the Lipschitz continuity of composite functions, an upper bound on the Lipschitz constant of a DNN~\eqref{eqn:dnn_definition} with 1-Lipschitz activation functions is the product of the spectral norms, or equivalently, of the maximum singular values of the weight matrices:
\begin{equation}
\label{eqn:upper_bound_main}
 \overline{\rho}(f(\bx)) = \prod_{l=1}^{L+1} ||\mathbf{W}_{l}||_2,
\end{equation}
where $ \overline{\rho}(f(\bx))$ denotes the upper bound on the Lipschitz constant of the DNN, $ ||\mathbf{W}_{l}||_2$ denotes the spectral norm or the maximum singular value of the weight matrix~$\mathbf{W}_{l}$. As derived in \cite{combettes2019lipschitz}, a lower bound $ \underline{\rho}(f(\bx))$ on the Lipschitz constant of a DNN  is
\begin{equation}
\label{eqn:lower_bound_main}
 \underline{\rho}(f(\bx)) = ||\mathbf{W}_{L+1}\mathbf{W}_{L}\cdots \mathbf{W}_{1}||_2,
\end{equation}
which corresponds to the Lipschitz constant of a purely linear network (i.e., a network with activation nonlinearities removed).

Note that the upper and lower bounds on the Lipschitz constant of a DNN in~\eqref{eqn:upper_bound_main} and \eqref{eqn:lower_bound_main} depend only on the maximum singular values of the weight matrices and their product. In the following analysis, we leverage random matrix theory to derive expressions of the bounds in~\eqref{eqn:upper_bound_main} and \eqref{eqn:lower_bound_main} in terms of the width and depth of the DNN and the variance of the weight parameters~$\sigma_w^2$.

\subsection{Estimates of the Lipschitz Constant Bounds Based on Extreme Singular Value Theorem}
In this subsection, we establish a connection between the Lipschitz constant of a DNN and its architecture (i.e., width and depth) based on the extreme singular value theory for random matrices.

\subsubsection{Upper Bound}
In this part, we show that, for a sufficiently large $\sigma_w$, the expected upper bound on the Lipschitz constant~\eqref{eqn:upper_bound_main} and hence the attainable expressiveness of a DNN increases exponentially with depth and polynomially with width. 
To start our discussion, we state the following result from random matrix theory on the extreme singular values of Gaussian random matrices:
\begin{theorem}[Gaussian Random Matrix~\citep{rudelson2010non}]
Let $\mathbf{A}$ be an $(N \times n)$ matrix whose elements are independent standard normal random variables. Then, $\sqrt{N}-\sqrt{n}\le\mathbb{E}[\lambda_\text{min}(\mathbf{A})] \le \mathbb{E}[\lambda_\text{max}(\mathbf{A})]\le \sqrt{N}+\sqrt{n}$, where $\lambda_\text{min}$ and $\lambda_\text{max}$ denote the minimum and maximum singular values of $\mathbf{A}$, respectively, and $\mathbb{E}[\cdot]$ represents the expected value.
\label{thm:extreme_singular_value}
\end{theorem}
Note that, for a Gaussian random matrix, the theorem above allows us to infer the extreme singular values of the matrix without explicitly knowing the values of its elements. By representing the weight parameters of a DNN as i.i.d. Gaussian random variables, we can leverage this result to estimate the upper bound of the Lipschitz constant~\eqref{eqn:upper_bound_main}. In particular, by applying Theorem~\ref{thm:extreme_singular_value}, we prove the following theorem in App.~\ref{app:upper_bound}:
\begin{theorem}[Upper Bound on Lipschitz Constant of a Gaussian Random DNN]
Consider a DNN defined in~\eqref{eqn:dnn_definition}, where the weight parameters are independent Gaussian random variables distributed as~$\mathcal{N}(0,\sigma_w^2)$ with $\sigma_w^2$ denoting the variance of the Gaussian distribution, and where the activation functions are 1-Lipschitz. The expected Lipschitz constant of the DNN is upper bounded by $\prod_{l=1}^{L+1} \sigma_w\left(\sqrt{n_l}+ \sqrt{n_{l-1}}\right)$.
\label{theorem:upper_bound}
\end{theorem}
Theorem~\ref{theorem:upper_bound} allows us to obtain an intuition about the expected attainable Lipschitz constant and thus the flexibility of a DNN as we vary its width $n_l$ for $l = 1,..., L$ and depth $L+1$. To compare to established results~\cite{serra2017bounding,raghu2017expressive}, we set the width of the hidden layers to $n$ (i.e., $n_l = n$ for $l = 1,...,L$), then the expected Lipschitz constant of a DNN with Gaussian random weights is upper bounded by $O\left((2\sigma_w)^{L+1} n^{\frac{L+1}{2}}\right)$. For $\sigma_w \ge \frac{1}{2\sqrt{n}}$, this upper bound increases exponentially with depth and polynomially with width. This observation is consistent with the results on the expressiveness measured by the number of linear regions for piecewise linear networks~\cite{serra2017bounding,raghu2017expressive} and the expressiveness measured by the trajectory length for Gaussian random networks~\cite{raghu2017expressive}. 

\subsubsection{Lower Bound}
\label{subsubsec:lower_bound}
Similarly based on the extreme singular value theorem for random matrices, we present a conjecture on the lower bound of the Lipschitz constant~\eqref{eqn:lower_bound_main}. We include a justification of the conjecture in App.~\ref{app:lower_bound} and empirically illustrate the result in Sec.~\ref{sec:numerical_examples}.
\begin{conjecture}[Lower Bound on Lipschitz Constant of a Gaussian Random DNN] Consider a DNN defined in~\eqref{eqn:dnn_definition} where the weight parameters are independent Gaussian random variables distributed as~$\mathcal{N}(0,\sigma_w^2)$ and the activation functions are 1-Lipschitz. The Lipschitz constant of the DNN is approximately lower bounded by $\left(\sigma_w^{L+1} \prod_{l=1}^{L}\sqrt{n_l}\right)\left(\sqrt{n_{L+1}} +\sqrt{n_0}+O(\sqrt{n_0})\right)$. 
\label{theorem:lower_bound}
\end{conjecture}

Based on Conjecture~\ref{theorem:lower_bound}, if we consider a DNN with constant width $n$ (i.e., $n_l = n$ for $l = 1,...,L$), the Lipschitz constant of the DNN with independent Gaussian weight parameters is approximately lower bounded by $\Omega\left(\sigma_w^{L+1}n^{\frac{L}{2}}\right)$, which also increases exponentially in depth and polynomially in the width of the DNN given sufficiently large $\sigma_w$ (i.e., $\sigma_w\ge \frac{1}{\sqrt{n}}$). Interestingly, we note that, for the case where $n\gg 1$ and $L\gg 1$, this asymptotic lower bound based on the Lipschitz constant of the DNN coincides with the expressiveness lower bound based on the output trajectory length measure for DNNs with ReLU activation functions~\cite{raghu2017expressive}. This connection is sensible since the expressiveness measure in~\cite{raghu2017expressive} can be intuitively thought of as the extent to which the DNN stretches a trajectory in its input space, which is a property related to the Lipschitz constant of a DNN (see App.~\ref{app:connection} for further details).\\[-1em]

Note that, for both the upper and lower bound analysis, we require the magnitude of $\sigma_w$ to be sufficiently large. Intuitively, a small $\sigma_w$ means that the magnitude of the weights are small. In the extreme case, where all weights are zero, a deep architecture cannot be expressive in any notion of expressiveness (e.g., number of linear regions). We therefore require the spread of the weights~$\sigma_w$ to be sufficiently large to exploit the expressivity of the deep layers. This lower bound is typically not restrictive; as an example, $1/\sqrt{n}$ is approximately 0.22 for $n = 20$.

\subsubsection{Differences Compared to Other Expressiveness Measures}
In this work, we propose to use the Lipschitz constant of a DNN as a measure of its expressiveness. In contrast to existing expressiveness measures, a Lipschitz-based characterization has two benefits: 
\begin{itemize}
\item \textit{Less assumptions on the DNN:} As compared to previous work on piecewise linear DNNs~\cite{pascanu2013number,montufar2014number,arora2016understanding,serra2017bounding}, by considering the Lipschitz constant as the expressiveness measure, we do not constrain ourself to DNNs with specific activation functions such as ReLUs or hard tanh. In our analysis, we only require the activation function to be 1-Lipschitz, which is satisfied by most commonly used activations that include but are not limited to ReLU, tanh, hard tanh, and sigmoid. 
\item \textit{Towards understanding DNN expressiveness for practical applications:} In contrast to expressiveness measures such as the number of linear regions~\cite{pascanu2013number,montufar2014number,arora2016understanding,serra2017bounding} and trajectory length~\cite{raghu2017expressive}, the Lipschitz constant is a generic property for Lipschitz continuous nonlinear functions. For regression problems, the expressiveness characterization through the Lipschitz constant allows us to make sensible comparisons between a DNN and the function it approximates. For control applications, the Lipschitz constant also plays a critical role in stability analysis. The Lipschitz-based characterization of the expressiveness of a DNN has the potential to facilitate the design of deep architectures for safe and efficient learning in a closed-loop control setup.
\end{itemize}

\section{Numerical Examples}
\label{sec:numerical_examples}
In this section, we provide numerical examples that illustrate the insights on the expressiveness of DNNs based on the results in Sec.~\ref{sec:lipschitz_constant}. In particular, we show the connection between the architectural properties of a DNN and its expressiveness.

\subsection{Bounds on the Lipschitz Constant of a DNN}
\begin{figure}
\centering
\includegraphics[width=0.42\textwidth]{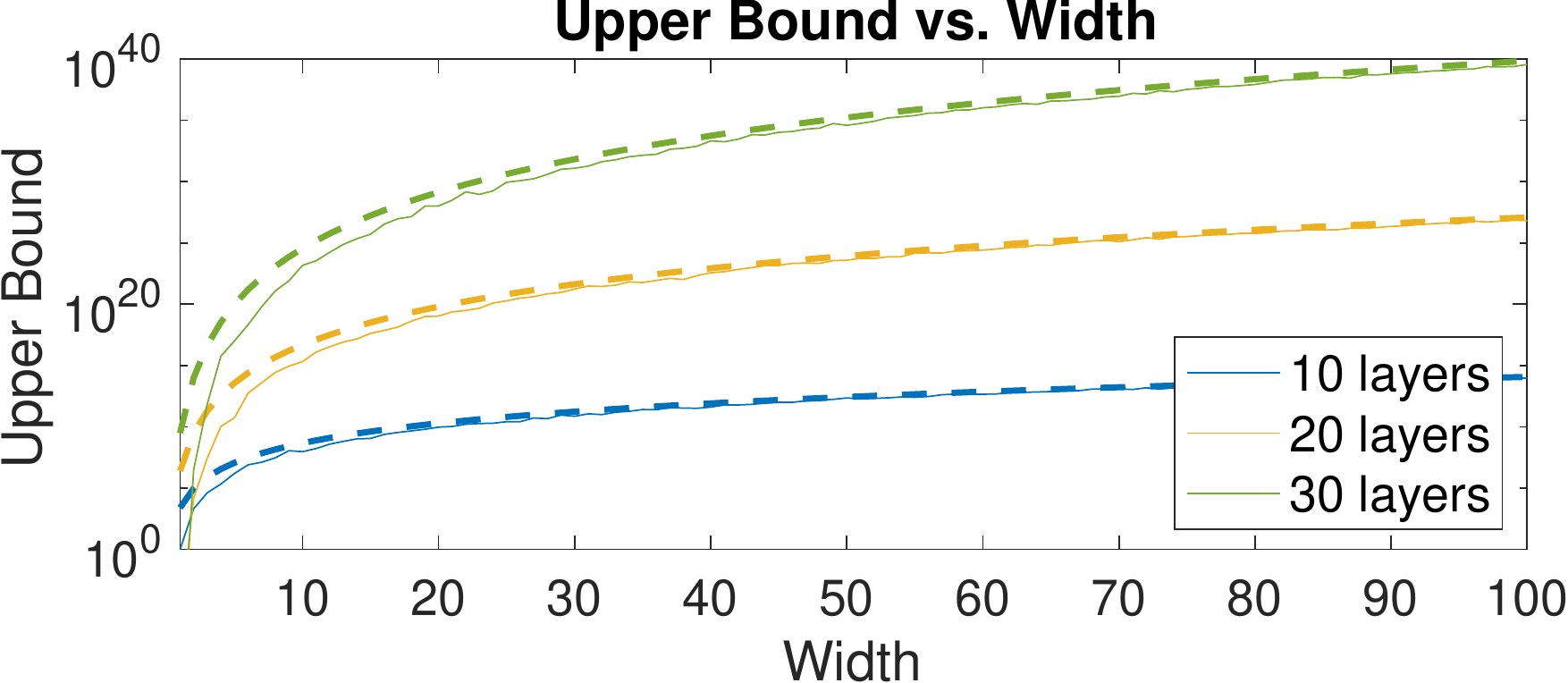}\hspace{1em}
\includegraphics[width=0.42\textwidth]{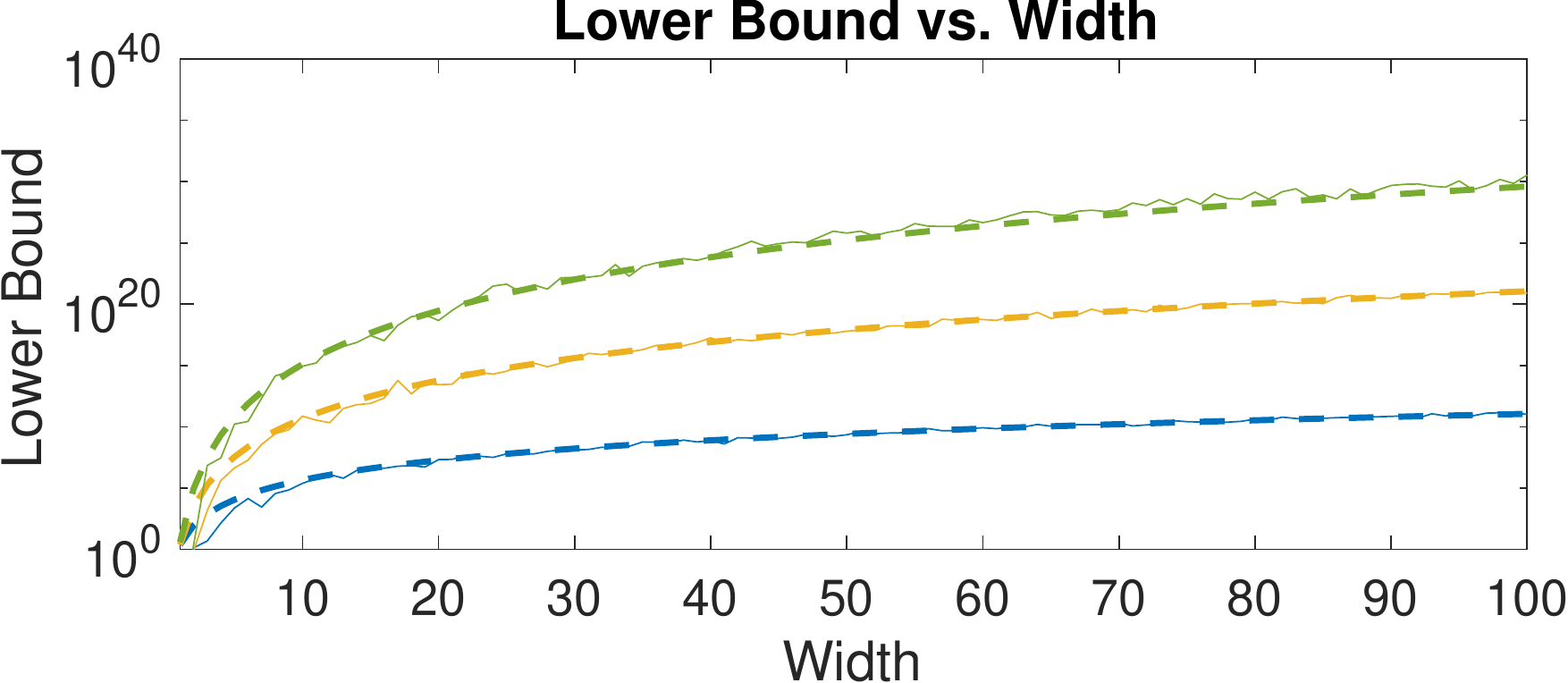}\\[0.5em]
\includegraphics[width=0.42\textwidth]{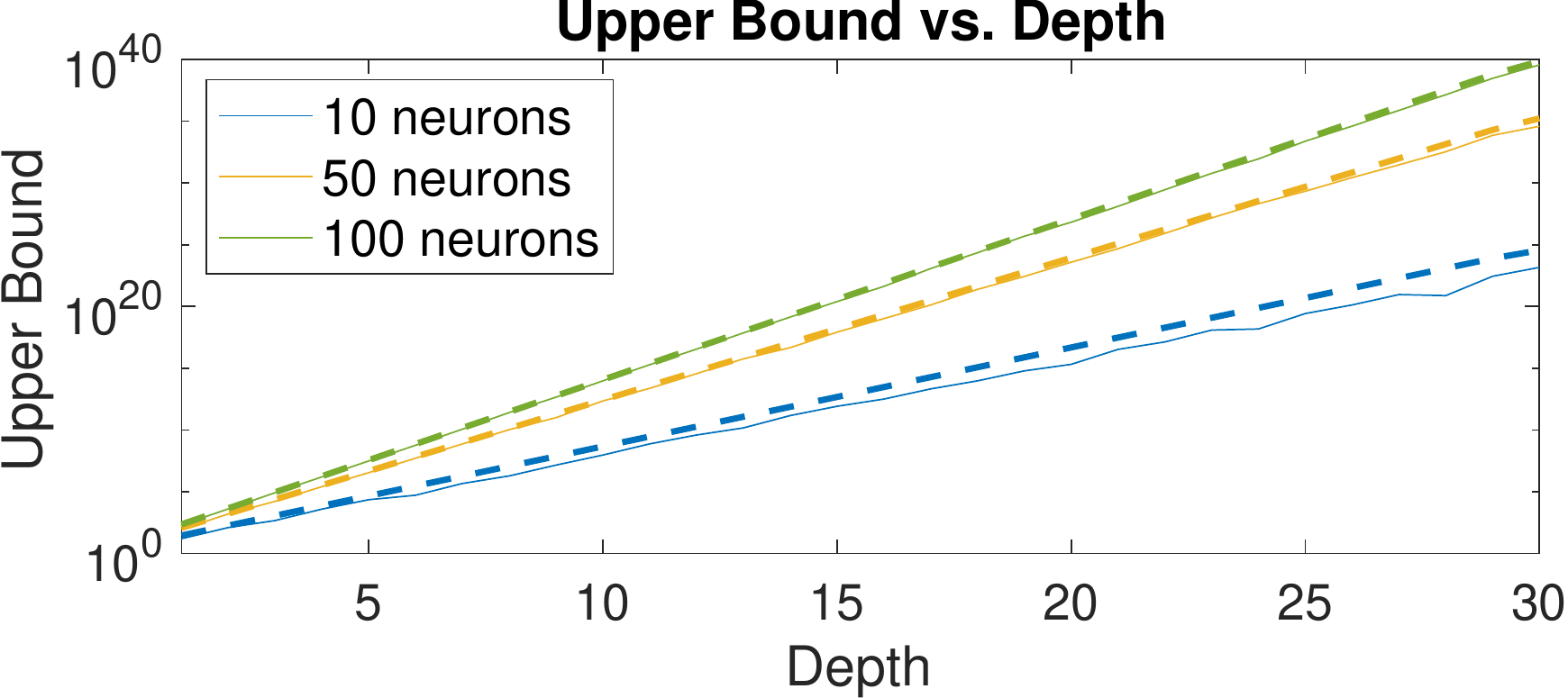}\hspace{1em}
\includegraphics[width=0.42\textwidth]{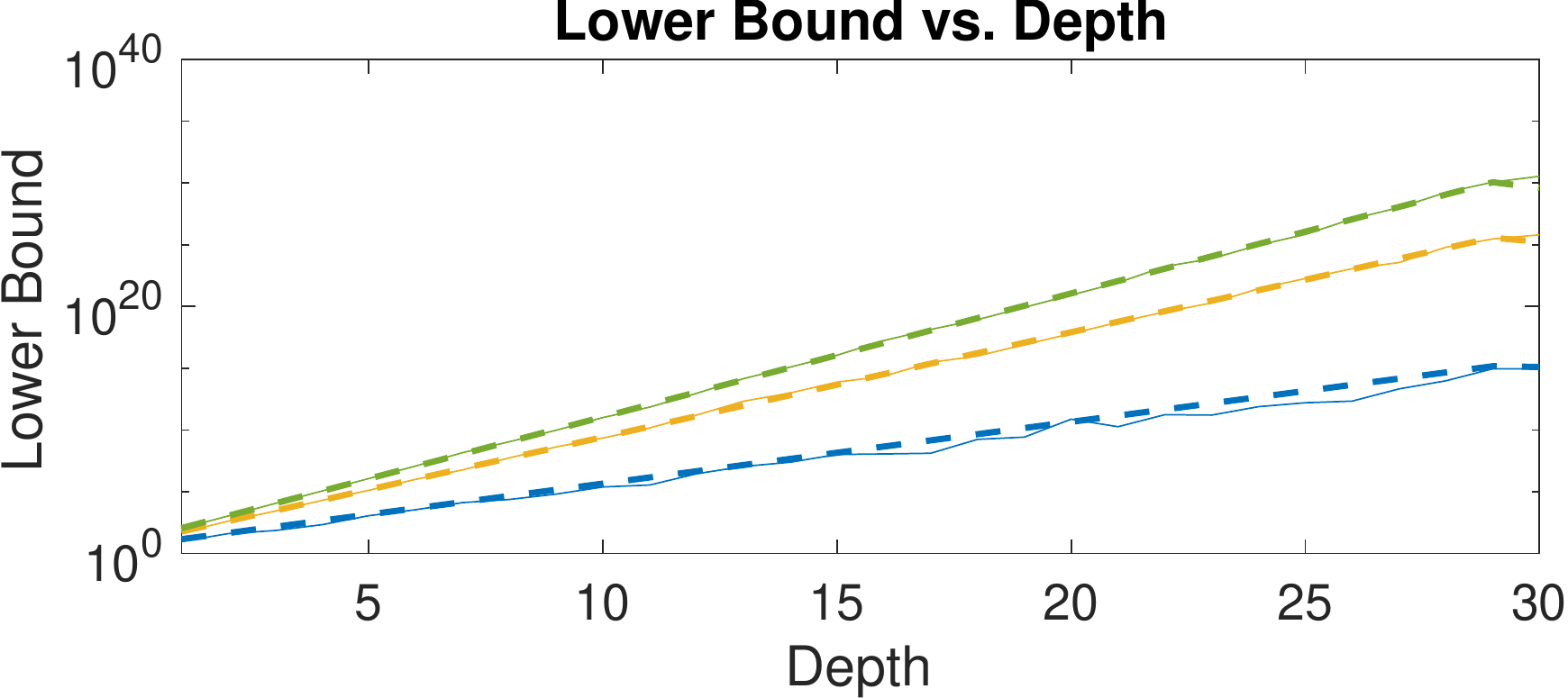}
\caption{A comparison of the estimated lower and upper bounds of the Lipschitz constant of DNNs based on Theorem~\ref{theorem:upper_bound} and Conjecture~\ref{theorem:lower_bound} \textit{(dashed lines)}, and the lower and upper bounds computed based on~\eqref{eqn:upper_bound_main} and~\eqref{eqn:lower_bound_main} with the actual weight values \textit{(solid lines)}.}
\label{fig:est_vs_act}
\vspace{-1em}
\end{figure}

To visualize the results of Sec.~\eqref{sec:lipschitz_constant}, we randomly sample the weight parameters of DNNs from a zero-mean, unit variance Gaussian distribution and compare the upper and lower bounds on the Lipschitz constants of these DNNs as we increase its width and depth. To examine the quality of the estimated Lipschitz constant bounds from Sec.~\eqref{sec:lipschitz_constant}, we show a comparison of the estimated  bounds computed based on Theorem~\ref{theorem:upper_bound} and Conjecture~\ref{theorem:lower_bound} and the bounds computed directly based on~\eqref{eqn:upper_bound_main} and \eqref{eqn:lower_bound_main} in Fig.~\ref{fig:est_vs_act}. From these plots, we see that there is a close correspondence between the Lipschitz constant bounds computed based on Theorem~\ref{theorem:upper_bound} and Conjecture~\ref{theorem:lower_bound}, which assumes random matrices, and the bounds computed based on~\eqref{eqn:upper_bound_main} and \eqref{eqn:lower_bound_main} based on the actual network weights. This result verifies that the bounds provided in Theorem~\ref{theorem:upper_bound} and Conjecture~\ref{theorem:lower_bound} are good approximations of the bounds on the Lipschitz constant of a fixed DNN based on~\eqref{eqn:upper_bound_main} and \eqref{eqn:lower_bound_main}.  We note that here we compute the bounds in~\eqref{eqn:upper_bound_main} and \eqref{eqn:lower_bound_main} directly based on the sampled weight parameters that are known for this simulation study; in general, to understand the implications of a DNN architecture based on Theorem~\ref{theorem:upper_bound} and Conjecture~\ref{theorem:lower_bound}, we do not rely on knowing the weights explicitly.

\begin{figure}
\centering
\includegraphics[width=0.45\textwidth]{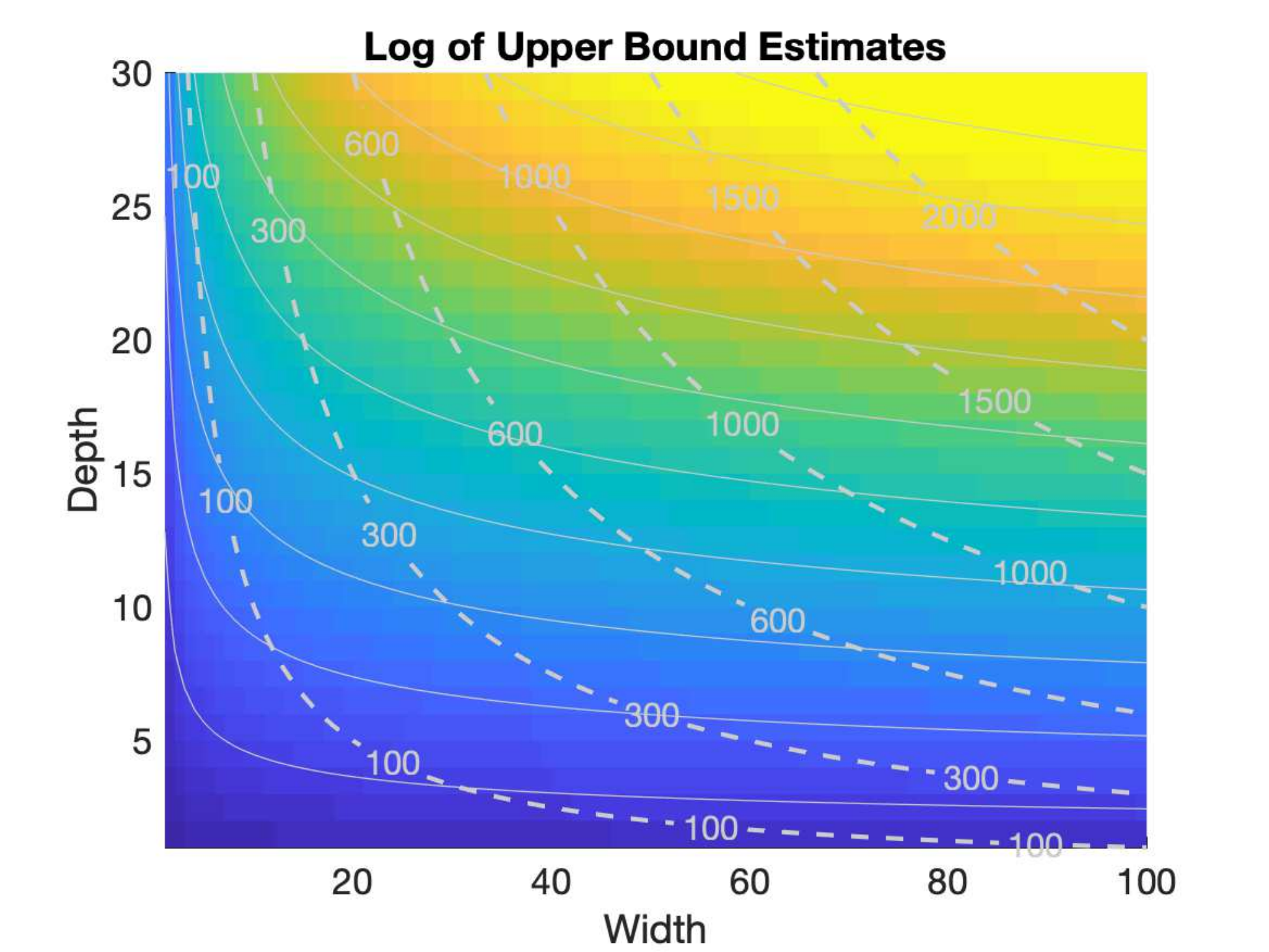}\hspace{-1em}
\includegraphics[width=0.45\textwidth]{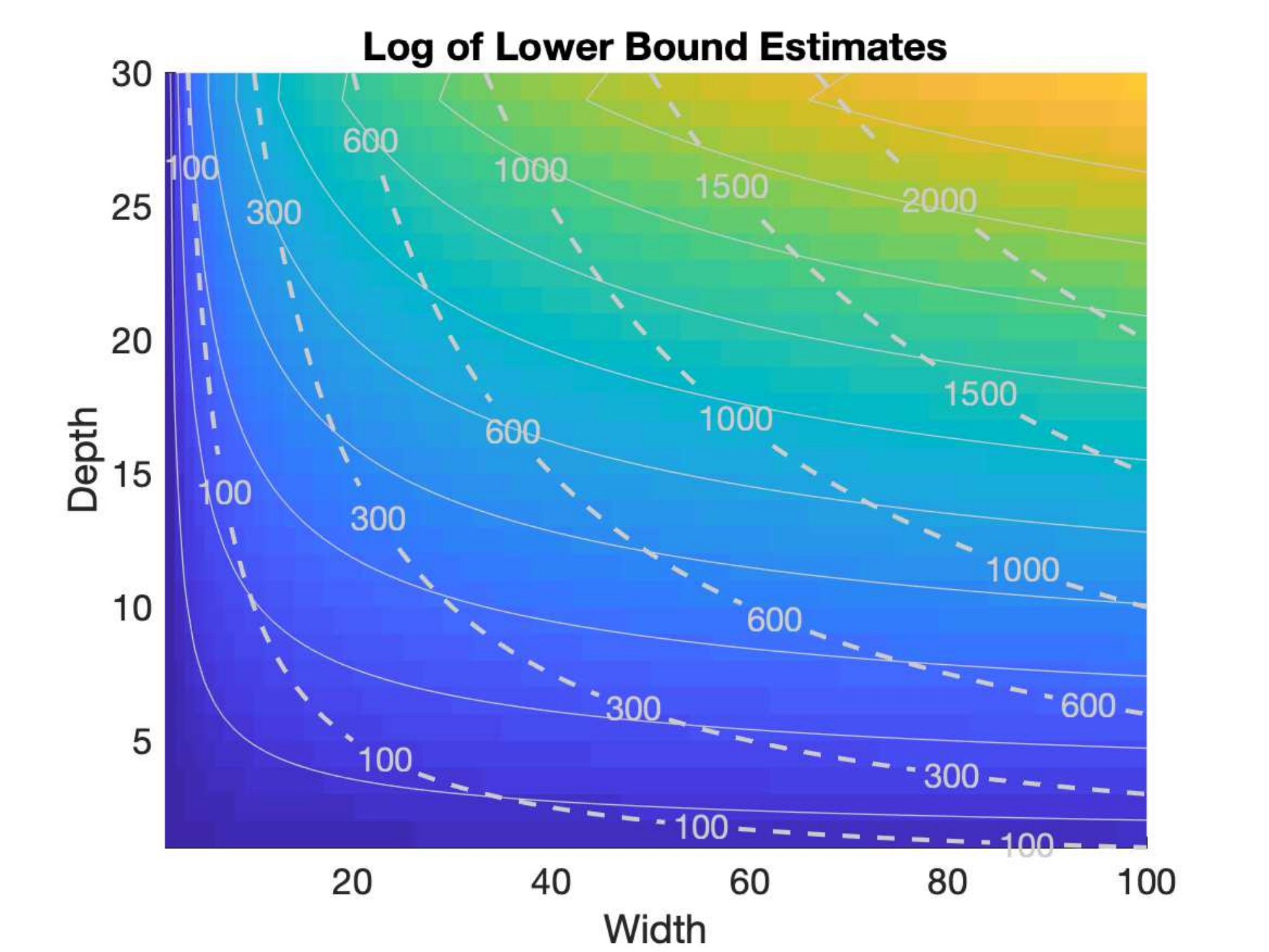}
\includegraphics[trim={16.6cm 0 0 0},clip,width=0.071\textwidth]{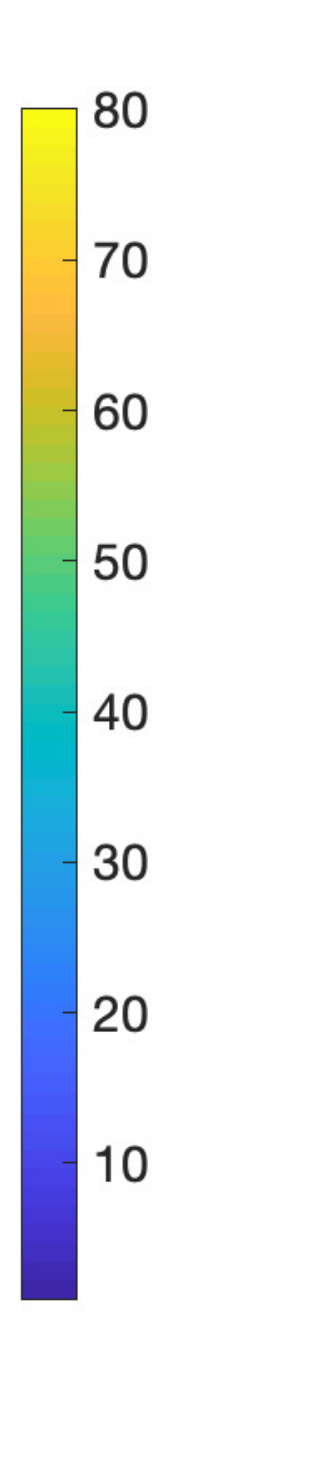}
\caption{Estimated lower and upper bounds on the Lipschitz constant of DNNs with varying widths and depths. The growth of the lower and upper bound is relatively faster in depth than in width. The dashed contour lines correspond to DNN architectures with equal numbers of neurons, and the solid contour lines correspond to levels of constant Lipschitz bounds.}
\label{fig:sim_est_bounds}
\vspace{-1em}
\end{figure}

Figure~\ref{fig:sim_est_bounds} shows the upper and lower bounds of the Lipschitz constant based on Theorem~\ref{theorem:upper_bound} and Conjecture~\ref{theorem:lower_bound} for different DNN architectures. By inspecting horizontal slices and vertical slices of the plots in Fig.~\ref{fig:sim_est_bounds}, which correspond to the top and bottom plots in Fig.~\ref{fig:est_vs_act}, we see that the upper and lower bounds of the Lipschitz constant of a DNN increase exponentially with depth and polynomially with width. The dashed contour lines in the plots show DNN architectures with the same number of neurons. As we trace one of the contour lines from left to right, we see that increasing width and decreasing depth reduces the bounds of the Lipschitz constants, which indicates a decrease in the expressiveness of the deep architecture. Similar to the discussion in~\cite{montufar2017number}, based on our formulation, we also see that, given the same number of neurons, deeper networks are more compact representations of nonlinear functions.

\subsection{Towards Learning Deep Models for Control}
To illustrate the implication of the expressiveness of a DNN for control, we consider a simple system setup and examine the stability of the system when we use a DNN with different architectures in the loop. In particular, we consider a system that is represented by
\begin{equation}
\dot{\bx} = \mathbf{A}\bx + f(\bx),
\label{eqn:sim_simple_system}
\end{equation}
where $\bx$ is the state, $\mathbf{A}$ is Hurwitz, and $ f(\bx)$ is a function parametrized by a DNN. By Lyapunov's direct method, one can show that a condition that guarantees stability of the system~\eqref{eqn:sim_simple_system} is 
\begin{equation}
\rho(f(\bx))\le \lambda_\text{min}(\mathbf{Q})/\left(2 \lambda_\text{max} (\mathbf{P})\right),
\label{eqn:sim_lip_safe_upper_bound}
\end{equation}
where $\rho(f(\bx))$ denotes the Lipschitz constant of the DNN, $\mathbf{Q}$ is a positive definite matrix, $\mathbf{P}$ is the corresponding solution to the Lyapunov equation $PA + A^TP = -Q$, and $\lambda_\text{min}$ and $\lambda_\text{max}$ are the minimum and maximum eigenvalues of a matrix, respectively.

For an illustration, we set $\mathbf{A} = \begin{scriptsize}\begin{bmatrix}
           0     &   2700\\
       -3600 &      -5400
\end{bmatrix}\end{scriptsize}$.
We compare five DNN architectures with different widths and depths but the same number of neurons. For each DNN architecture, we sample 50 DNNs with i.i.d. zero-mean, unit variance Gaussian weight parameters. We note that out of the five architectures, we know based on Theorem~\ref{theorem:upper_bound} that the first case, a DNN with a hidden layer of 300 neurons, has an estimated upper bound on the Lipschitz constant less than the safe upper bound in~\eqref{eqn:sim_lip_safe_upper_bound}, and system~\eqref{eqn:sim_simple_system} is stable. In contrast, as we can see from Fig.~\ref{fig:sim_est_bounds}, when we decrease the width and increase the depth of a DNN, its Lipschitz constant increases and system~\eqref{eqn:sim_simple_system} is less likely to be stable. Table~\ref{tab:sim_stability_summary} shows empirical results for the relationship between the architectural properties of a DNN and the stability of the system. This means, in practice, one may want to carefully choose an appropriate DNN architecture, or, alternatively, regularize the weight parameters, to ensure stability of a learning-based control system. We consider our insights to be a step towards providing design guidelines for DNN architectures, for example, for closed-loop control applications.

\begin{table}[]
\centering
\caption{Likelihood of stable system with different DNN architectures.}
\vspace{-0.5em}
\footnotesize
\begin{tabular}{lccccc}
\hline\hline
Architecture (width $\times$ depth) & $300 \times 1$ & $100\times 3$ & $50\times 6$ & $20\times 15$ & 10$\times$ 30 \\\hline
Likelihood of stable system (\%)    & 100            & 100           & 40           & 32           & 32\\\hline\hline
\end{tabular}
\label{tab:sim_stability_summary}
\vspace{-3.5em}
\end{table}

\section{Discussion on the Assumption of Gaussian Random Weight Matrices}
\label{sec:random_weights}
In this work, we considered DNNs with Gaussian random weight matrices to facilitate analysis of their expressiveness. In this section, we examine if this assumption is reasonable for practical applications. In particular, we examine, through some examples, the accuracy of estimating the maximum singular value of the weight matrices based on Theorem~\ref{thm:extreme_singular_value} when the assumption of Gaussian random matrices does not hold exactly.

\begin{table}
\vspace{3em}
\caption{True and estimated maximum singular values of weight matrices in trained networks.}
\vspace{-0.5em}
\footnotesize
\centering
\begin{tabular}{c|cc|cc}
\hline\hline
&\multicolumn{2}{c|}{Network 1 (64 Neurons)} & \multicolumn{2}{c}{Network 2 (256 Neurons)} \\\hline
&True Norm   & Estimated Norm   & True Norm     & Estimated Norm     \\\hline
$\mathbf{W}_1$ & 36 & 38.5 & 197 & 208  \\
$\mathbf{W}_2$ & 7.38 & 8.31 &1.92 & 2.04     \\\hline\hline      
\end{tabular}
\label{tab:norm_estimate}
\vspace{-1em}
\end{table}

\begin{wrapfigure}{r}{4.5cm}
\centering
\vspace{-1.2em}
\includegraphics[trim=1.2cm 6.5cm 10cm 0.5cm, clip, width=4.5cm]{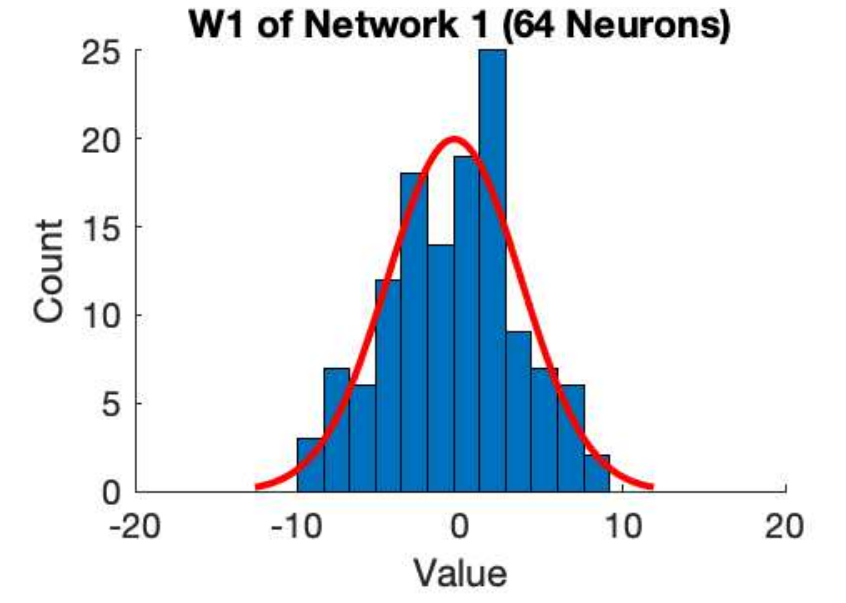}
\includegraphics[trim=1.2cm 6.5cm 10cm 0.5cm, clip, width=4.5cm]{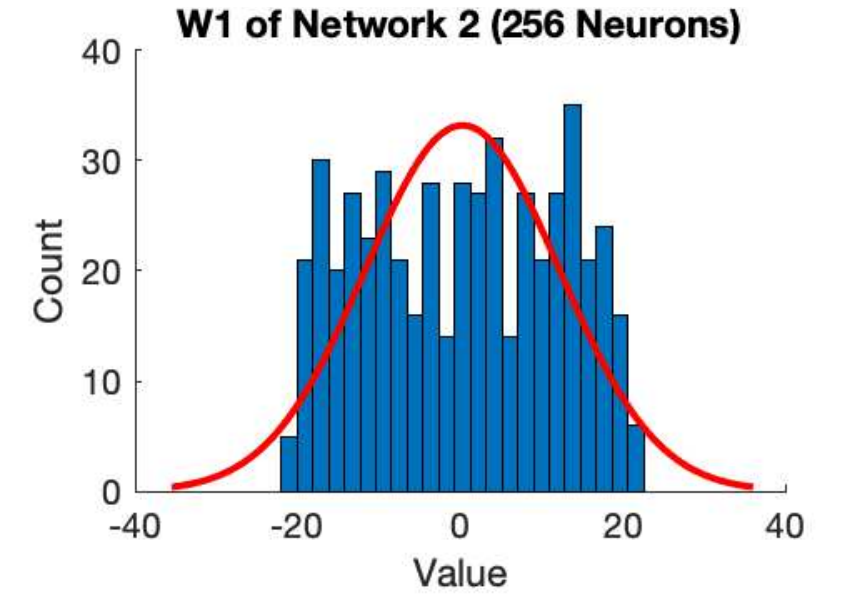}
\caption{Examples of weight distributions in trained networks. The red lines correspond to fitted Gaussian curves.
}
\label{fig:weight_distributions}
\end{wrapfigure}

To examine the properties of weight matrices in trained networks, we consider a regression problem. The true function to be approximated has two inputs and one output. Fig.~\ref{fig:weight_distributions} shows the distributions of two weight matrices from two trained networks with different architectures, and Table~\ref{tab:norm_estimate} summarizes their maximum singular values. By inspecting the distributions (Fig.~\ref{fig:weight_distributions}), we see that the weights are not necessarily always Gaussian-distributed; however, the estimates of the maximum singular values of the matrices based on the assumption of random weights are very close to the true maximum singular values (Table~\ref{tab:norm_estimate}). 
Based on Bai-Yin's law for extreme singular values of random matrices with more general distributions~\cite{rudelson2010non}, we can infer that the expected maximum singular value based on Theorem~\ref{thm:extreme_singular_value} is an approximation of the true maximum singular value of a random matrix with an error of $O(\sigma_w \sqrt{n})$, where $\sigma_w$ is the standard deviation of the weight distribution and $n$ is the matrix column dimension. In future, we plan to explore the properties of the weight matrices of trained networks and examine their relation to random matrix theory.

\section{Conclusion}
In this paper, we presented a new perspective on the expressiveness of DNNs based on their Lipschitz properties. Using random matrix theory, we showed that, given the spread of the weights is sufficiently large (i.e., $\sigma_w \ge \frac{1}{\sqrt{n_l}}$ for $l = 1,...,L$), the expressiveness of a DNN measured by its Lipschitz constant grows exponentially with depth and polynomially with width. This result is similar to the results based on other expressiveness measures discussed in the current literature. By considering the Lipschitz constant as a measure of DNN expressiveness, we can more sensibly understand the implication of being `deep' in the context of function approximation for applications including safe learning-based control.

\color{black}
\bibliography{reference}

\clearpage
\appendix
\section{Proofs of Main Results in Sec.~\ref{sec:lipschitz_constant}}
\subsection{Proof of Theorem~\ref{theorem:upper_bound}: Upper Bound on Lipschitz Constant of a Gaussian Random DNN}
\label{app:upper_bound}
The following is a proof for Theorem~\ref{theorem:upper_bound} presented in Sec.~\ref{sec:lipschitz_constant}. In the following proof, based on the extreme singular value theorem for random matrices (Theorem~\ref{thm:extreme_singular_value}), we derive an expression for the upper bound on the Lipschitz constant of a DNN in terms of its width and depth.

\vspace{1em}
\begin{proof}
Consider a random matrix  $\mathbf{A}\in\Rbb^{N\times n}$ whose elements are independent Gaussian random variables distributed as $\mathcal{N}(0,\sigma_w^2)$. As a result of Theorem~\ref{thm:extreme_singular_value} and the homogeneity of the matrix norm, the expected maximum singular value of $\mathbf{A}$ is upper bounded by $\mathbb{E}[\lambda_\text{max}(\mathbf{A})]\le \sigma_w(\sqrt{N}+\sqrt{n})$. By assumption, the elements of each weight matrix $\mathbf{W}_l$ are distributed as $\mathcal{N}(0,\sigma_w^2)$. The expected spectral norm, or equivalently the expected maximum singular value, of the weight matrices are upper bounded as follows:
\begin{equation}
\label{eqn:max_sv_weights}
\Ebb[||\mathbf{W}_l||_2] = \Ebb[\lambda_\text{max}(\mathbf{W}_l)]\le  \sigma_w(\sqrt{n_l}+\sqrt{n_{l-1}}).
\end{equation}
Since the weight matrices are independent, by substituting~\eqref{eqn:max_sv_weights} into~\eqref{eqn:upper_bound_main}, we have the following expected upper bound on the Lipschitz constant of the DNN:
\begin{equation}
\label{eqn:upper_bound_proof}
\Ebb[\overline{\rho}(f(\mathbf{x}))] = \prod_{l=1}^{L+1} \Ebb[||\mathbf{W}_{l}||_2]\le \prod_{l=1}^{L+1}\sigma_w \left(\sqrt{n_l}+ \sqrt{n_{l-1}}\right).
\end{equation}
The expression in~\eqref{eqn:upper_bound_proof} establishes a connection between the upper bound on the Lipschitz constant of a DNN and its architecture, which is represented by the dimensions of the weight matrices in this analysis. This result allows us to obtain insights on the expressiveness of a DNN without explicitly knowing the values of its weights. 
\end{proof}

\subsection{Justification of Conjecture~3: Lower Bound  on Lipschitz Constant of a Gaussian Random DNN}
\label{app:lower_bound}
To derive an estimate of the lower bound in~\eqref{eqn:lower_bound_main}, we first note that the product of random Gaussian matrices is in general not a Gaussian random matrix. In deriving the lower bound, we need to consider a more general class of matrices than in Theorem~\ref{thm:extreme_singular_value}:
\begin{theorem}[Random Matrix~\citep{rudelson2010non}]
Let $\mathbf{A}$ be an $N \times n$ matrix whose elements are independent random variables with zero mean, unit variance, and finite fourth moment. Suppose that the dimensions $N$ and $n$ grow to infinity with $N/n$ converging to a constant in $[0,1]$.
Then, $\mathbb{E}[\lambda_\text{min}(\mathbf{A})]  = \sqrt{N}-\sqrt{n} + O(\sqrt{n})$ and $ \mathbb{E}[\lambda_\text{max}(\mathbf{A})] = \sqrt{N}+\sqrt{n} + O(\sqrt{n})$ almost surely.
\label{thm:extreme_singular_value_general}
\end{theorem}
In contrast to Theorem~\ref{thm:extreme_singular_value}, the above theorem is applicable to a wider class of random matrices with independent elements; however, this result is an asymptotic result in the limit of sufficiently large $N$ and $n$. For practical DNNs where the dimensions of the weight matrices are sufficiently large, this theorem allows us to derive an approximate lower bound for~\eqref{eqn:lower_bound_main}. We provide a justification of Conjecture~\ref{theorem:lower_bound} presented in Sec.~\ref{sec:lipschitz_constant} of our manuscript below:

\begin{justification}
We consider two random matrices $\mathbf{A}_1\in\Rbb^{N\times n_1}$ and $\mathbf{A}_2\in\Rbb^{n_2\times N}$ whose elements are independent zero-mean random variables with variances $\sigma_{a1}^2$ and $\sigma_{a2}^2$, respectively. The $i$th row and $j$th column element of the matrix product $\mathbf{A}_{21} = \mathbf{A}_2\mathbf{A}_1$ is $\sum_{k=1}^{N} a_{2,ik}a_{1,kj}$, where $a_{1,kj}$ denotes the $k$th row and $j$th column element of $\mathbf{A}_1$ and $a_{2,ik}$ denotes the $i$th row and $k$th column element of~$\mathbf{A}_2$. Here, in our derivation, we make a conjecture that the elements of the product matrix of random matrices with elements being i.i.d. zero-mean random variables approximately preserve independence. Based on this conjecture, we derive an expression of the variance the elements of $\mathbf{A}_{21}$. Without loss of generality, we consider the $i$th row and $j$th column element of $\mathbf{A}_{21}$. Since, by assumption, the elements of $\mathbf{A}_1$ and $\mathbf{A}_2$ have zero mean and are i.i.d., the variance of the $i$th row and $j$th column element of $\mathbf{A}_{21}$ is
\begin{align}
\sigma_{21}^2 &= \mathbb{V}\left[\sum_{k=1}^{N} a_{2,ik}a_{1,kj} \right]=\sum_{k=1}^{N}\: \mathbb{V}\left[a_{2,ik}a_{1,kj}\right]=N \sigma_{a1}^2 \sigma_{a2}^2,
\end{align}
where $\mathbb{V}$ denotes the variance of a random variable, and $\mathbb{V}\left[a_{2,ik}a_{1,kj}\right]=\sigma_{a1}^2 \sigma_{a2}^2, \forall k= 1,2,...,N$ is the variance of the product of an element of $\mathbf{A}_1$ and an element of $\mathbf{A}_2$. The standard deviation of elements in the product of $\mathbf{A}_{21}$ can be written as
\begin{equation}
\label{eqn:variance_of_product}
\sigma_{21} = \sqrt{N} \sigma_{a1}\sigma_{a2}.
\end{equation}

By applying~\eqref{eqn:variance_of_product} recursively, we can derive an estimate of the bound in~\eqref{eqn:lower_bound_main}, which is the spectral norm of the product of random matrices. In particular, a recursive relationship in the standard deviations of the product of random matrices can be written as
\begin{equation}
\sigma_{w,1:l} = \sqrt{n_{l-1}}\sigma_{w,1:l-1}\sigma_{w},
\end{equation}
where $\sigma_{w,1:l} $ denotes the standard deviation of the product of random matrices $\mathbf{W}_l\mathbf{W}_{l-1}\cdots \mathbf{W}_1$. For the product random matrix $\mathbf{W}_{L+1}\mathbf{W}_{L}\cdots \mathbf{W}_1$ in~\eqref{eqn:lower_bound_main}, we have
\begin{equation}
\label{eqn:standard_deviation_prod}
\sigma_{1:L+1} = \sigma_w^{L+1}\prod_{l=1}^{L}\sqrt{n_l}.
\end{equation}

As above, we make a conjecture that the elements of the product matrix constructed from the random weight matrices $\mathbf{W}_1, \mathbf{W}_2,..., \mathbf{W}_{L+1}$ are independent. Since the elements of the product matrix are the sums of products of independent zero-mean random variables by construction, the elements of the product matrix have zero mean. Moreover, since the elements of the weight matrices $\mathbf{W}_1, \mathbf{W}_2,..., \mathbf{W}_{L+1}$ are assumed to be Gaussian distributed, they have finite fourth moments. Further by the properties of the sum and product of random variables~\citep{dufour2003properties}, the elements of the product matrix constructed from the weight matrices $\mathbf{W}_1, \mathbf{W}_2,..., \mathbf{W}_{L+1}$ also have finite fourth moments. By Theorem~\ref{thm:extreme_singular_value_general} and the homogeneity of matrix norms, a random matrix $\mathbf{M}$ whose elements are i.i.d. random variables with mean 0, variance $\sigma_w^2$, and finite fourth moment, the expected maximum singular value of $\mathbf{M}$ is given by
\begin{equation}
\label{eqn:expected_max_sv_variance}
\mathbb{E}[\lambda_\text{max}(\mathbf{M})] = \sigma_m\left(\sqrt{N}+\sqrt{n} + O(\sqrt{n})\right).
\end{equation}
Based on~\eqref{eqn:standard_deviation_prod} and~\eqref{eqn:expected_max_sv_variance}, an estimate of the expected lower bound of the Lipschitz constant in~\eqref{eqn:lower_bound_main} is 
\begin{equation}
\label{eqn:lower_bound_proof}
 \mathbb{E}[\underline{\rho}(f(\bx))] =\mathbb{E}[ ||\mathbf{W}_{L+1}\cdots \mathbf{W}_{1}||_2 ]= \left(\sigma_w^{L+1} \prod_{l=1}^{L}\sqrt{n_l}\right) \left(\sqrt{n_{L+1}} +\sqrt{n_0}+ O(\sqrt{n_0})\right).
\end{equation}
Similar to the upper bound, this expected lower bound on the Lipschitz constant allows us to infer the Lipschitz constant of a DNN based on its architectural properties.
\end{justification}
\vspace{1em}

In Sec.~\ref{sec:numerical_examples} of the manuscript, we empirically show that the expression in~\eqref{eqn:lower_bound_proof} is a reasonable approximation of the lower bound of the Lipschitz constant of a DNN in~\eqref{eqn:lower_bound_main}. However, we note that, in our justification above, we make an assumption that the elements of the product matrix constructed from random matrices whose elements are i.i.d. zero-mean Gaussian random variables preserve independence. This is a conjecture that requires further investigation. We would like to further look into results on multiplications of random matrices to improve this result. 

\section{Connection to the Result Based on Output Trajectory Length}
\label{app:connection}
In this appendix, we show a connection between our result and the result in~\cite{raghu2017expressive}. Both our work and~\cite{raghu2017expressive} consider DNNs with i.i.d. zero-mean Gaussian weight parameters. In our work, we use the Lipschitz constant as a measure of the expressiveness of a DNN, while in~\cite{raghu2017expressive}, the proposed expressiveness measure of a DNN is the expected length of an output trajectory as the DNN traverses a one-dimensional trajectory in its input space. Intuitively, as an input trajectory is passed through a DNN, it is deformed by the linear weight layers and the nonlinear activation layers; the output trajectory length measure in~\cite{raghu2017expressive} is the extent to which the DNN `stretches' a trajectory given in the input space.

By considering the expected output trajectory length as the expressiveness measure,~\cite{raghu2017expressive} prove the following result:

\begin{theorem}[Lower Bound on Output Trajectory Length~\citep{raghu2017expressive}]
Let $f(\bx)$ be a DNN with ReLU activation functions and weights being i.i.d. Gaussian random variables $\mathcal{N}(0,\sigma_w^2)$, and let $\bx(t)$ be a one-dimensional trajectory with $\bx(t+\delta)$ having a non-trivial perpendicular component to $\bx(t)$ for all $t,\delta$. Denote $\mathbf{h}_l(\bx(t)) = \mathbf{h}_l(t)$ as the image of the trajectory in the $l$th layer of the DNN. 
The expected output trajectory length of the DNN is lower bounded by
\begin{equation}
\Ebb[\eta(\mathbf{h}_{L+1}(t))] \ge O\left( \frac{\sigma_w n}{\sqrt{n+1}}\right)^{L+1} \eta(\bx(t)),
\label{eqn:lower_bound_trajectory_length}
\end{equation}
where $\eta(\bx(t))= \bigintsss_t \left\vert\left\vert\frac{d\bx(t)}{dt}\right\vert\right\vert_2 dt$ is the trajectory length and $n$ is the width of the DNN.
\end{theorem}
\begin{figure}
\centering
\includegraphics[width=8cm]{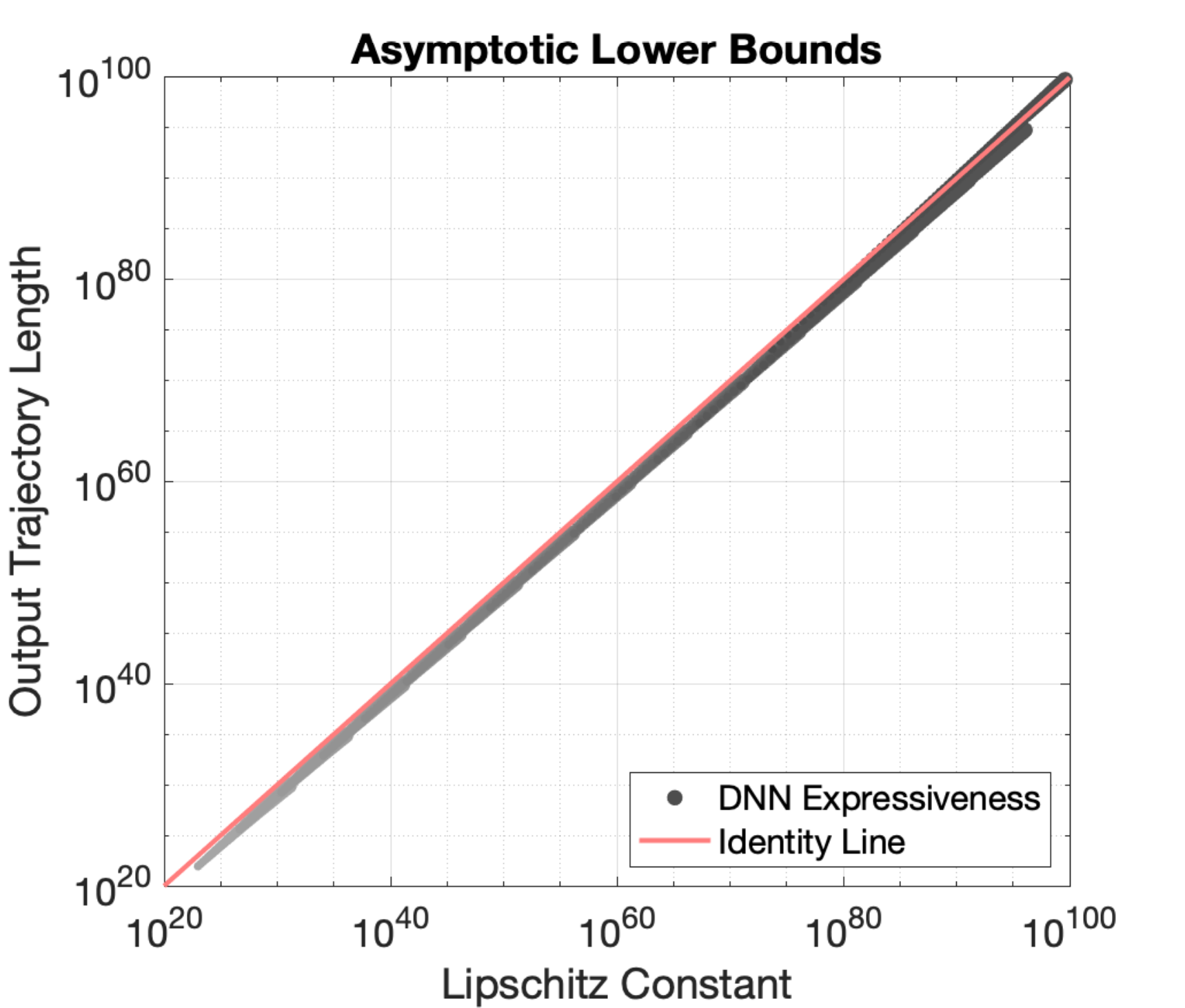}
\caption{Expressiveness of a ReLU DNN measured by output trajectory length~\cite{raghu2017expressive} versus the expressiveness by the proposed Lipschitz constant approach.  The dots in grey correspond to the calculated expressiveness measures for DNNs with different widths and depths, and the red solid line is the identity line. The result is generated for DNNs with width and depth ranging between 30 and 100, respectively.}
\label{fig:trajlength_vs_lipschitz}
\end{figure} 

Note that, if we consider the expected output trajectory length normalized by the input trajectory length (i.e., the `stretch' of the trajectory), we can establish a connection with the lower bound in~\eqref{eqn:lower_bound_trajectory_length} and the lower bound we derived based on Lipschitz constant expressiveness characterization in Sec.~\ref{subsubsec:lower_bound}. In particular, in Sec.~\ref{subsubsec:lower_bound}, we showed that for a DNN with a constant width $n$ (i.e., $n_l = n$ for $l = 1,...,L$), the asymptotic lower bound on the Lipschitz constant of the DNN is $O\left( \sigma_w^{L+1} n^{\frac{L}{2}}\right)$. On the other hand, the normalized lower bound on the expected output trajectory in~\eqref{eqn:lower_bound_trajectory_length} can be written as $O\left( \sigma_w^{L+1}\left(\frac{n}{\sqrt{n+1}}\right)^{L+1}\right)$. For $n\gg 1$ and $L\gg 1$, this asymptotic lower bound from~\eqref{eqn:lower_bound_trajectory_length} coincides with the asymptotic lower bound we obtained based on the Lipschitz constant measure of expressiveness. 
Fig.~\ref{fig:trajlength_vs_lipschitz} illustrates this connection between our proposed expressiveness measure based on the Lipschitz constant of a DNN and the expressiveness measure based on the output trajectory length~\cite{raghu2017expressive} for a set of ReLU DNNs with different widths and depths. From the plot, we see that, for DNNs with different architectures, the correlation between the asymptotic lower bounds based on these two measures of expressiveness (grey dots) approximately coincides with the  identity line (red line). 

The observed connection between the two measures of expressiveness of a DNN is sensible. If we consider the input trajectory to a DNN to be represented by a set of discrete points, the length of the output trajectory captures the extent of `stretch' between pairs of points as they are passed through the DNN. Mathematically, the extent of `stretch' or the distance between two points in a DNN's output space in relation to the distance between the corresponding points in the input space is characterized by the Lipschitz property of the DNN. 

\end{document}